\documentclass[letterpaper, 10 pt, conference]{ieeeconf}  

\IEEEoverridecommandlockouts                              
                                                          
\usepackage[utf8]{inputenc}
\usepackage[T1]{fontenc}
\usepackage[pdftex]{graphicx}
\usepackage{graphicx}
\usepackage{tikz}
\usepackage{subfigure} 
\graphicspath{{../Figures/}}
\usepackage{booktabs} 
\usepackage{multirow} 
\usepackage{threeparttable} 
\usepackage{float}
\usepackage{multicol}
\usepackage{color, colortbl}
\usepackage{amsmath} 
\usepackage{array}
\usepackage[ruled,vlined,linesnumbered]{algorithm2e}
\usepackage{mathtools, nccmath} 
   
\usepackage{bm} 

\usepackage[symbol]{footmisc}

\title{\LARGE \bf
A Novel Probabilistic V2X Data Fusion Framework for \\Cooperative Perception
}

\author{Mao Shan, Karan Narula, Stewart Worrall, Yung Fei Wong, Julie Stephany Berrio Perez, \\Paul Gray and Eduardo Nebot 
\thanks{M. Shan, K. Narula, S. Worrall, J.S. Berrio Perez and E. Nebot are with the Australian Centre for Field Robotics, The University of Sydney, NSW 2006, Australia. E-mails: {\tt \{mao.shan, karan.narula, stewart.worrall, julie.berrioperez, eduardo.nebot\}@sydney.edu.au}}
\thanks{Y.F. Wong and P. Gray are with Cohda Wireless, 27 Greenhill Road, Wayville, SA 5034, Australia. E-mails: {\tt \{ricky.wong, paul.gray\}@cohdawireless.com}}
\thanks{This research is funded by iMOVE CRC and supported by the Cooperative Research Centres program, an Australian Government initiative.}
}

\begin{document}

\maketitle
\thispagestyle{empty}
\pagestyle{empty}

\begin{abstract}
The paper addresses the vehicle-to-X (V2X) data fusion for cooperative or collective perception (CP). This emerging and promising intelligent transportation systems (ITS) technology has enormous potential for improving efficiency and safety of road transportation. Recent advances in V2X communication primarily address the definition of V2X messages and data dissemination amongst ITS stations (ITS-Ss) in a traffic environment. Yet, a largely unsolved problem is how a connected vehicle (CV) can efficiently and consistently fuse its local perception information with the data received from other ITS-Ss. In this paper, we present a novel data fusion framework to fuse the local and V2X perception data for CP that considers the presence of cross-correlation. The proposed approach is validated through comprehensive results obtained from numerical simulation, CARLA simulation, and real-world experimentation that incorporates V2X-enabled intelligent platforms. The real-world experiment includes a CV, a connected and automated vehicle (CAV), and an intelligent roadside unit (IRSU) retrofitted with vision and lidar sensors. We also demonstrate how the fused CP information can improve the awareness of vulnerable road users (VRU) for CV/CAV, and how this information can be considered in path planning/decision making within the CAV to facilitate safe interactions.
\end{abstract}

\section{Introduction}
Recent years have witnessed increasing popularity of \textit{vehicle-to-X} (V2X) technology among researchers in the field of \textit{intelligent transportation systems} (ITS) and with automobile manufacturers, as it enables a vehicle to share essential information with other road users in a V2X network. The \textit{collective perception} (CP) service is currently being standardised by major regulatory organisations such as \textit{European Telecommunications Standard Institute} (ETSI), SAE and IEEE. The CP service enables an \textit{ITS station} (ITS-S) such as a \textit{connected vehicle} (CV) or an \textit{intelligent roadside unit} (IRSU) to share its perception information with adjacent ITS-Ss by exchanging \textit{Collective Perception Messages} (CPMs) via V2X communication. The ETSI specified CPMs convey abstract representations of perceived objects instead of raw sensory data, facilitating the interoperability between ITS-Ss of different types and from different manufactures.

The limitations of the onboard perception sensors of smart vehicles justify CP service's use for improving road safety. Despite significant advances in sensor technology, the perception capability of these local sensors is ultimately bounded in range and \textit{field of view} (FOV) due to their physical constraints. Besides, occluding objects in urban traffic environments such as buildings, trees, and other road users impose challenges in perception. In this case, a CV can benefit from the CP service in terms of improved awareness of surrounding road users, essential for ensuring road safety. Specifically, it facilitates a CV to extend its sensing range and improve sensing quality, redundancy, and robustness through sensor fusion across multiple ITS-Ss. Besides, the enhanced perception quality resulting from the cross-platform sensor fusion potentially relaxes the accuracy and reliability requirements of onboard perception sensors.

Most of the existing data fusion approaches for V2X based CP in the literature assume perceived information received from other ITS-Ss to be independent object observations for the convenience of ignoring estimate dependency in the fusion process. \cite{paper:RauchKlanner2012} presents a high-level object fusion framework for CP based on Ko-PER \cite{web:KoFAS} specified CPMs, where the perception information received from other ITS-Ss is modelled as a virtual sensor in its data fusion architecture.  \cite{paper:RauchMaier2013} investigates the inter-vehicle object association for facilitating data fusion for CP. Both methods treat the received perception data as pre-processed but independent sensory measurements so it does not require track-wise fusion. \cite{paper:MouawadMannoni2021_2} proposes a low complexity fusion scheme for CP. It yet again does not explicitly address the track-to-track fusion and their strategy is complicated by considering a handful of conditional branches in the deterministic association of the local and V2X objects.

A high-level object fusion framework for V2X is also presented in \cite{paper:GuntherMennenga2016}. Instead of completing data fusion in a joint fashion, it separates the fusion process of local sensor data and the perceived information in received \textit{Environmental Perception Messages} (EPMs). It benefits from not mixing up local and V2X perception by maintaining two lists of tracked objects. However, this approach does not fully utilise the advantages of the CP. The tracking quality of locally perceived objects will not be improved by the perception data from other ITS-Ss, which would otherwise be achieved in joint fusion schemes. There are also occupancy-grid based data fusion methods for CP, such as in \cite{paper:MouawadMannoni2021_1, paper:GodoyJimenez}. This kind of fusion framework is considered computationally expensive, where pixel-wise iterative fusion is required.

Arguably, sharing and fusing only independent sensor observations for V2X-based CP means only CVs with sensing capability can contribute. Besides, the dissemination of perception information in this case is limited to a single-hop range. It also causes repeated processing of the same perception data across multiple receiving CVs, which could be avoided if object tracks, which are post-fusion information, are shared in V2X communication. Combining tracks from multiple sources in a CP system is not straightforward. It requires further attention to handling the common prior information from a single platform and across different platforms. The simplest example is when two communicating nodes keep exchanging and fusing data under the wrong assumption of information independence. In this scenario, the covariance matrix, which indicates the uncertainty level of the fusion result, would be reduced when it should have remained the same, as identical pieces of information are incorrectly duplicated in the data fusion process.

The problem is more complex in a V2X-based CP setting, where the tracks of perceived objects shared amongst ITS-Ss are dependent on previous estimates and become cross-correlated as they are typically exchanged between arbitrary pairs of CVs in the V2X network. To tackle the problem of common prior information, one can choose to keep track of communication history and broadcast it in the network for identifying the common prior information. In this case, customised data fields must be added along with the communicated packets, which is infeasible to achieve with the current ETSI CPM specification. Alternatively, using the \textit{covariance intersection} (CI) algorithm \cite{paper:JulierUhlmann1997}, consistent estimation results are produced in cross-platform data fusion with unknown cross-correlation. The work in \cite{paper:AlligWanielik2019ITSC} proposes a variant of Ko-PER CPM and analyses the trade-off between message size resulted from enabling optional data fields in the CPM and global fusion accuracy. It performs CI-based track-to-track fusion relying on object ID association, and the method is only evaluated in a numerical simulation with two CVs.

The first and main contribution of the paper is a novel V2X data fusion framework for CP which can support the joint fusion of different types of perception information in a host CV, including its local perception data, independent object observations and object tracks it receives from other cooperative ITS-Ss through V2X communication. Specifically, a CI-based probabilistic track-to-track fusion scheme is proposed to achieve safe and consistent fusion of perception information from multiple V2X sources without requiring knowledge of any cross-correlation. The proposed data fusion framework is validated through comprehensive qualitative and quantitative results obtained from numerical simulation, CARLA simulator, and real-world experimentation with V2X-enabled intelligent platforms. 

The second contribution of the paper lies in the demonstration of the critical role the CP plays in the safety of CAV operations, in particular, in scenarios where a CAV is unable to detect other road users using local sensors due to visual occlusion or perception sensor failures. It is also demonstrated how the CAV considered the fused V2X perception information in its path planning and decision making to facilitate safe interaction with \textit{vulnerable road users} (VRU).

\section{The Proposed Framework}
\label{sec:framework}

\subsection{Overview}

\begin{figure}[t]
	\centering
	\includegraphics[width=2.4in, trim={5.4in 2.5in 1.55in 0.95in}, clip]{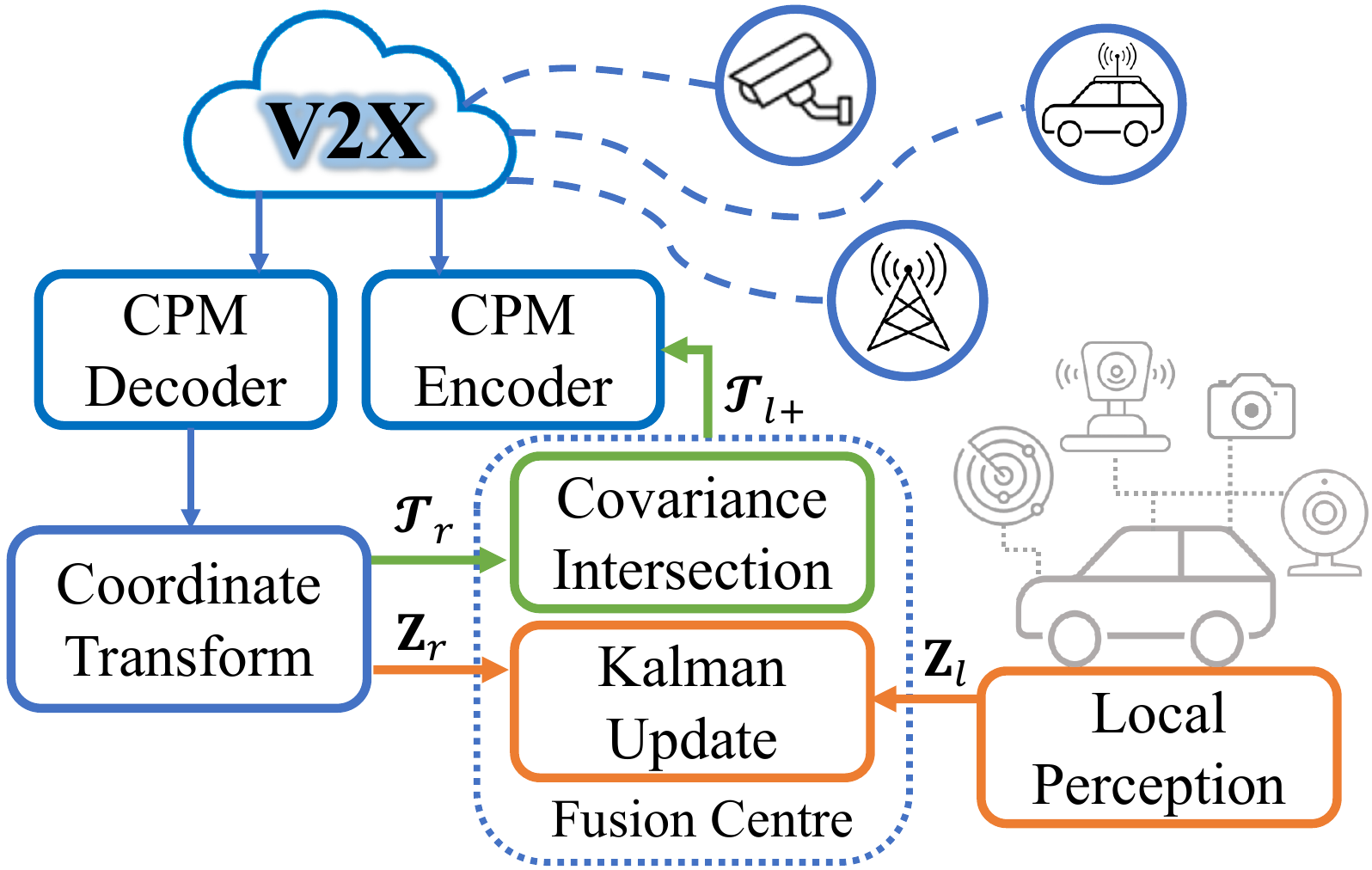}
	\caption{An overview of the proposed V2X data fusion scheme for CVs. It shows how the remote perception information received in the form of ETSI CPMs from multiple other ITS-Ss through V2X communication is processed by the fusion centre jointly with the local perception data within a CV. The updated object tracks as a result of data fusion are then encoded into CPMs before being transmitted to other ITS-Ss.}
	\label{fig:cp_fusion_scheme}
\end{figure}

Figure \ref{fig:cp_fusion_scheme} presents an overview of the proposed V2X data fusion framework for CVs. The framework starts with decoding the CPMs received from other ITS-Ss, such as other CVs and special infrastructure, via specialised V2X communication devices. Each ETSI CPM consists of an ITS PDU header and five types of information containers accommodating mandatory and optional \textit{data elements} (DEs) and \textit{data frames} (DFs), which are:
\begin{enumerate}
\item A \textit{CPM Management Container}, which indicates station type, such as a vehicle or an IRSU, and reference position of the transmitting ITS-S.
\item An optional \textit{Station Data Container}, which provides additional information about the originating station. This includes the heading, speed, and dimensions when the originating station is a vehicle.
\item Optional \textit{Sensor Information Containers}, which describe the type and specifications of the equipped sensors of the transmitting ITS-S, including sensor IDs, types, and detection areas.
\item Optional \textit{Perceived Object Containers}, each of which describes the dynamics and properties of a perceived object, such as type, position, speed, heading, and dimensions. These perceived object descriptions are registered in the coordinate system of the originating station.
\item Optional \textit{Free Space Addendum Containers}, which describe different confidence levels for certain areas within the sensor detection areas.
\end{enumerate}

When perceived object information is received and decoded, it is not usable to the receiving CV until it is transformed into its local coordinate system. We stress that uncertainty bounds associated with the perception information, which have been represented in \textit{Perceived Object Container} of the CPM specification, are indispensable in successive data fusion and thus have to be considered. Besides, it is essential for the coordinate transformation also to incorporate the uncertainty contained in the pose estimation of both ITS-Ss. The originating ITS-S information including its pose with associated uncertainty is encapsulated within the \textit{CPM Management Container} and \textit{Station Data Container} in the CPM definition. Consequently, the transformed object state estimates contain uncertainty that is mainly a combined result of uncertainty in the localisation of ITS-Ss and sensory perception. Interested readers can refer to our prior work in \cite{paper:ShanNarula2021} for the details of the handling of CPMs and coordinate transformation of the perceived object information.

Following the coordinate transformation with uncertainty, the perceived objects information from remote sources are fused into the fusion centre within the receiving CV. The CPM generation rules support different abstraction levels for perceived object descriptions for the implementation flexibility, which can derive from low-level abstract observations made by individual sensors or the results of the high-level data fusion, such as object tracking results. Both types of perception information can co-exist in the received CPMs, depending on the fusion configuration of the transmitting ITS-Ss. Therefore, both types of perceived object information are supported, and processed differently in the proposed data fusion framework.

The low-level object observations are denoted as $\bf{Z}_{r}$, which are considered independent observations generated by onboard perception sensors in the originating ITS-S. Fusing these independent sensor observations in the fusion centre is primarily the same to the fusion of the local perception sensory data, denoted as $\bf{Z}_{l}$, using a traditional Bayesian filtering paradigm, such as a Kalman filter. The paper will not elaborate on the fusion of independent observations, which is extensively covered in the literature, but instead focus on fusing the object tracks contained in received CPMs, denoted as $\bm{\mathcal{T}}_{r}$, with the local object tracks, denoted as $\bm{\mathcal{T}}_{l}$. It will require consistent track-to-track fusion on the receiving ITS-S side. In this case, special care needs to be taken to avoid duplicating common prior information, which causes inconsistency in state estimation. The developed CP framework employs the CI algorithm for fusing tracks information from two or more sources without knowing the cross-correlation information between them. Details on the CI and the proposed track-to-track fusion approach can be found in Section \ref{sec:track2track_fusion}.

As the implementation of fusion centre in this paper, a variant of \textit{Gaussian mixture probability hypothesis density} (GMPHD) filter \cite{paper:VoMa2006} is employed to track multiple road users within the receiving CV. The GMPHD-based tracker is considered attractive due to its inherent convenience in handling track initiation, track termination, probabilistic data association, and clutter. Compared with the naive GMPHD algorithm, the particular variant implemented in the paper is improved with measurement-driven initiation of new tracks, track identity management, and track-to-track fusion. We however stress that the proposed V2X data fusion framework is intentionally designed to be generic such that one can choose to use a road user tracking algorithm of their preference. Yet, the proposed track-to-track fusion approach is considered an essential add-on component to the existing road user tracker to be able to fuse remote tracks received from other ITS-Ss.

Lastly, after the fusion with local sensory data and remote perceived objects information, the updated object tracks $\bm{\mathcal{T}}_{l+}$ are then encoded into CPMs before being transmitted to other ITS-Ss through V2X communication.

\subsection{Track-to-Track Fusion}
\label{sec:track2track_fusion}

A traditional Kalman-based filter only works under the independent information assumption, while the CI algorithm can consistently fuse estimates, in particular, in the case without the knowledge of existing cross-correlation between them. We employ this strategy for the track-to-track fusion for received object tracks in the proposed V2X data fusion framework for CP.

\subsubsection{Covariance Intersection}

Assume we would like to combine two Gaussian estimates \(\mathcal{N}\left(\textbf{a},\textbf{A}\right)\) and \(\mathcal{N}\left(\textbf{b},\textbf{B}\right)\) without knowing their cross-correlation. In this work, we consider the CI formulation generalised for the case where the two estimates can possess different dimensions, as in \cite{paper:ArambelRago2001}, which means\useshortskip
\begin{equation}
\label{eq:obs_matrix}
    dim\left(\textbf{b}\right) = dim\left(\textbf{H} \textbf{a}\right),
\end{equation}
where \(\textbf{H}\) is defined as an observation matrix.

The fused estimate \(\mathcal{N}\left(\textbf{c},\textbf{C}\right)\), which will have the same dimension as \(\mathcal{N}\left(\textbf{a},\textbf{A}\right)\), is then computed using the CI algorithm as\useshortskip
\begin{equation}
    \mathcal{N}\left(\textbf{c},\textbf{C}\right) = CI\left(\mathcal{N}\left(\textbf{a},\textbf{A}\right), \mathcal{N}\left(\textbf{b},\textbf{B}\right)\right),
\end{equation}
where\useshortskip
\begin{equation}
\label{eq:CI}
\begin{split}
    \textbf{C}^{-1} &= \omega \textbf{A}^{-1} + \left(1-\omega\right) \textbf{H}^{T} \textbf{B}^{-1} \textbf{H}\\
    \textbf{c} &= \textbf{C} \left(\omega \textbf{A}^{-1} \textbf{a} + \left(1-\omega\right) \textbf{H}^{T} \textbf{B}^{-1} \textbf{b}\right).
\end{split}
\end{equation}
In \eqref{eq:CI}, \(\omega \in \left[0, 1\right]\) is a scalar obtained by solving the optimisation problem of minimising the trace or determinant of \(\textbf{C}\). For the case of minimising the determinant, it can be written as\useshortskip
\begin{equation}
    \omega = \underset{\omega}{\arg\min} det\left(\textbf{C}\right).
\end{equation}

\begin{figure}[!t]
	\centering
	\includegraphics[width=2.4in]{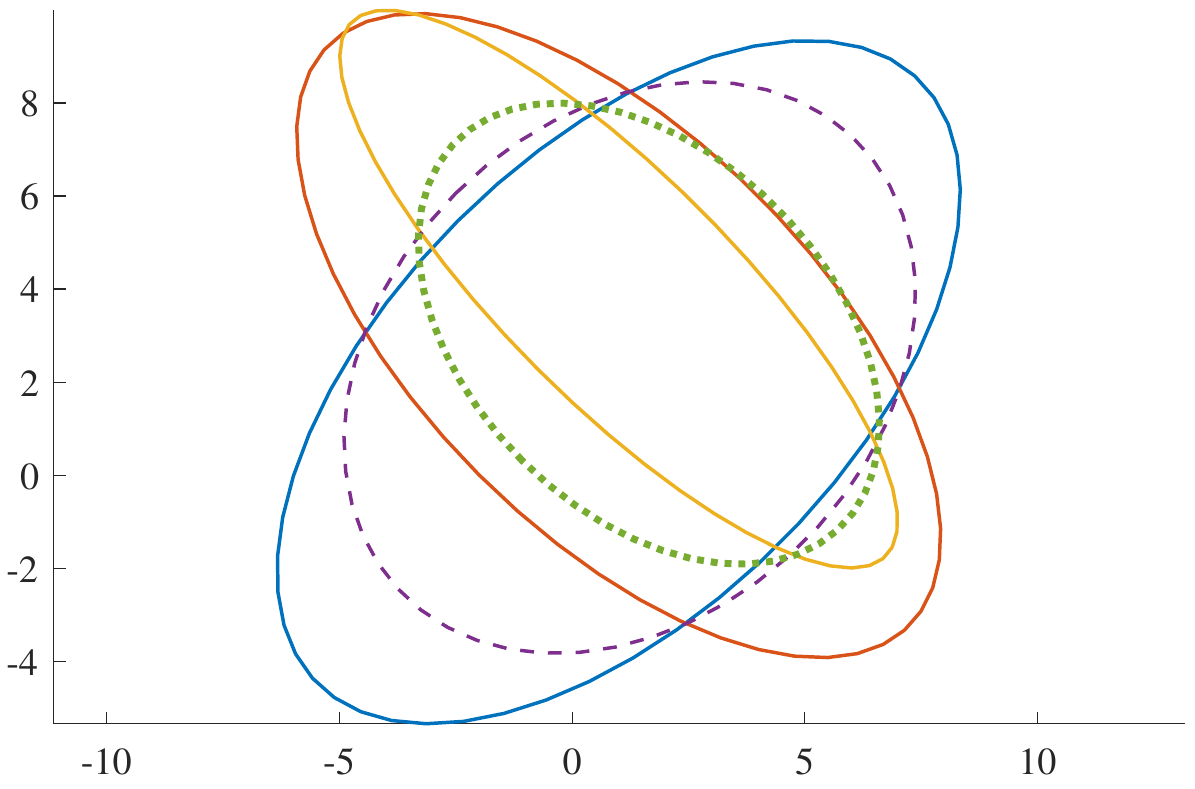}
	\caption{Illustration of fusing Gaussian estimates from three sources using the CI algorithm. For the convenience of illustration, the three estimates considered are visualised in 2D as blue, red, and yellow solid ellipses, representing their respective 95\% confidence ellipses. The CI first fuses the two estimates in blue and red with \(\omega = 0.791\) and yields the fused estimate shown in 2D as the purple dashed ellipse. The fused estimate and the third estimate in yellow are then fused using CI with \(\omega = 0.683\) this time to produce the estimate visualised as the green dotted ellipse.}
	\label{fig:ExampleCI}
\end{figure}

Figure \ref{fig:ExampleCI} shows a numerical example that fuses Gaussian estimates from multiple sources. It can be easily calculated that the fused estimate produced by CI has a smaller covariance than any estimate before fusion. Although not shown in the figure, the fused estimate remains the same when we fuse the result with the third estimate recursively for 10 more times, and \(\omega = 1\). This demonstrates one of the attractive properties of CI algorithm, which is the immunity to information duplicity. When two input estimates are truly independent, the CI fused estimates are generally more conservative than Kalman filtered results. However, CI is the safer option when the correlation of information from different sources is unclear.

\subsubsection{Tracklet Fusion Management}

The section will focus on the tracklet fusion management in the track-to-track fusion. The host CV can communicate with multiple other cooperative ITS-Ss at a time in a V2X setting. To keep the mathematical notation simple, the fusion management presented here covers the generic case between a host CV and another ITS-S. The very same fusion can be applied to every other cooperative ITS-S iteratively. Assume a CPM that the host CV receives from another ITS-S contains \(N\) remote object tracks, in the meantime, the host CV maintains \(M\) local object tracks. It is assumed that the track-to-track fusion happens at time step \(t\), therefore all the notations presented in the section has \(t\) dropped for simplicity. The collection of remote object tracks are denoted as \(\bm{\mathcal{T}}_{r} = \left\{T_{r}^{j}\right\}_{j=1}^{N}\), where each \(T_{r}\) is a tuple that can be decomposed to\useshortskip
\begin{equation}
\label{eq:remote_tracks}
    T_{r} = \left\{\bar{\textbf{x}}_{r}, \bm{\Sigma}_{r}, s_{r}, d_{r}\right\},
\end{equation}
where \(\bar{\textbf{x}}_{r}\) is track mean vector, \(\bm{\Sigma}_{r}\) is covariance matrix, \(s_{r}\) refers to the unique ID of the transmitting ITS-S, and \(d_{r}\) represents the track ID within the originating ITS-S. These elements are encoded into a CPM from the transmitter side.

Similarly, the set of local tracks are represented by \(\bm{\mathcal{T}}_{l} = \left\{T_{l}^{i}\right\}_{i=1}^{M}\), and each\useshortskip
\begin{equation}
\label{eq:local_tracks}
    T_{l} = \left\{\bar{\textbf{x}}_{l}, \bm{\Sigma}_{l}, w, d_{l}, \textbf{R}\right\},
\end{equation}
where \(w\) is the associated weight for the track, \(d_{l}\) represents the track ID within the host ITS-S, and the alias ID list \(\textbf{R} = \left\{\begin{bmatrix}s_{r}^{j}&d_{r}^{j}\end{bmatrix}\right\}_{j=1}^{P}\) is introduced. Each pair in the list refers to the alias ID that the local track \(T_{l}\) is known as by another ITS-S, and the very same track can exist in \(P\) other ITS-Ss. Initially, the list is empty, i.e., \(P = 0\). Details of the alias ID list will be given later. Unlike that in the remote track, the ID of the host ITS-S is not required in the local track. Also note that in many scenarios the state vector of a remote track can be a subset of a local track maintained in the tracker. Hence, the CI formulation presented in \eqref{eq:CI} is used to accommodate the case of unmatched length of state vectors of the remote and local tracks. Like in \eqref{eq:obs_matrix}, there exists an observation matrix \(\textbf{H}\) for the tracks, such that
\(
    dim\left(\bar{\textbf{x}}_{r}\right) = dim\left(\textbf{H} \bar{\textbf{x}}_{l}\right).
\)

As shown in \eqref{eq:remote_tracks} and \eqref{eq:local_tracks}, a track contains state estimates and its assigned track ID within the originating ITS-S, and different ITS-S have their own ways of initiating and managing track IDs. The same object in the real world may be known by different track IDs within different ITS-Ss. When an ITS-S receives a remote track, it would be impossible to determine if it matches up with a local track only by its track ID. To handle this issue, we introduce a simple yet effective track matching and fusion mechanism, which divides the track-to-track fusion into three subcases detailed in the following. 

\begin{itemize}
\item Given each remote track $T_{r} \in \bm{\mathcal{T}}_{r}$, the mechanism first performs fusion only for its any matched local track $T_{l} \in \bm{\mathcal{T}}_{l}$ if it searches the alias ID list $\textbf{R} \in T_{l}$, finding that $T_{l}$ has been fused with information from the particular cooperative ITS-S before, i.e., $\begin{bmatrix}s_{r}&d_{r}\end{bmatrix}$ exists in $\textbf{R}$. This deterministic association of remote tracks with the matched local tracks essentially reduces the computation overhead in track-to-track fusion.
\item For the remaining unmatched local tracks, they are fused only with those remote tracks received from the originating ITS-S for the first time. Specifically, it is determined based on the condition that $s_{r} \in T_{r}$ is not found in $\textbf{R} \in T_{l}$, and probabilistic track association is performed for each possible pair of unmatched local and remote tracks. Once a local track is fused with a remote track, $\begin{bmatrix}s_{r}&d_{r}\end{bmatrix}$ is added into $\textbf{R}$ for matching check at future time steps. It is also considered in this subcase the initiation of possible new local tracks based on the unmatched remote tracks.
\item Lastly, the group of local tracks not detected by the cooperative ITS-S are updated given their detection probability \(p_{D,r}\left(\cdot\right)\), as explained later in \eqref{eq:ci_update}.
\end{itemize}

Given a pair of $T_{l}$ and $T_{r}$ that are intended for fusion, the CI based track-to-track fusion is divided into two steps:
\begin{enumerate}
    \item Solving the optimal \(\omega \in \left(0, 1\right)\) for the CI algorithm:\useshortskip
\begin{equation}
\label{eq:ci_omega}
    \omega = \underset{\omega}{\arg\min} det\left(\bm{\Sigma}_{l\#}^{i}\right),
\end{equation}
where
\(\left(\bm{\Sigma}_{l\#}\right)^{-1} = \omega \left(\bm{\Sigma}_{l}\right)^{-1} + \left(1-\omega\right) \textbf{H}^{T} \left(\bm{\Sigma}_{r}\right)^{-1} \textbf{H}\).

    \item Estimate update:\useshortskip
\begin{equation}
\label{eq:ci_update}
\begin{split}
    \bar{\textbf{x}}_{l+} &= \bar{\textbf{x}}_{l} + \textbf{K} \textbf{y}\\
    \bm{\Sigma}_{l+} &= \left(\textbf{I} - \textbf{K}\textbf{H}\right) \bm{\Sigma}_{l*} \left(\textbf{I} - \textbf{K}\textbf{H}\right)^{T} + \textbf{K} \bm{\Sigma}_{r*} \textbf{K}^{T}\\
    w_{+} &= \alpha\  p_{D,r}\left(\bar{\textbf{x}}_{l}\right) q\left(T_{l},T_{r}\right) w,
\end{split}
\end{equation}
where\useshortskip
\begin{align*}
    \bm{\Sigma}_{l*} &= \frac{\bm{\Sigma}_{l}}{\omega} & \bm{\Sigma}_{r*} &= \frac{\bm{\Sigma}_{r}}{1-\omega} &
    \textbf{y} &= \bar{\textbf{x}}_{r} - \textbf{H} \bar{\textbf{x}}_{l}
\end{align*}\useshortskip
\begin{align*}
    \textbf{S} &= \bm{\Sigma}_{r*} + \textbf{H} \bm{\Sigma}_{l*} \textbf{H}^{T} &
    \textbf{K} &= \bm{\Sigma}_{l*} \textbf{H}^{T} \textbf{S}^{-1}
\end{align*}\useshortskip
\begin{align*}
q\left(T_{l},T_{r}\right) = \mathcal{N}\left(\textbf{y}, \textbf{S}\right),
\end{align*}
$\alpha$ is the normalising constant, and lastly, \(p_{D,r}\left(\bar{\textbf{x}}_{l}\right)\) is the detection probability of the cooperative ITS-S detecting a local target given its state mean vector \(\bar{\textbf{x}}_{l}\).
\end{enumerate}

The above \eqref{eq:ci_omega} and \eqref{eq:ci_update} can be combined as a function denoted as $\left\{\bar{\textbf{x}}_{l+}, \bm{\Sigma}_{l+}, w_{+}\right\} = CI\left(T_{l}, T_{r}\right)$. The pseudocode for the track matching and fusion algorithm can be found in Algorithm \ref{table:track2track}. Finally, the set of posterior local tracks, defined as $\bm{\mathcal{T}}_{l+}$, is a union of different subsets of unmatched and updated local tracks:\useshortskip
\begin{equation}
    \bm{\mathcal{T}}_{l+} = \bm{\mathcal{T}}_{l+}^{ud} \bigcup \bm{\mathcal{T}}_{l+}^{mt} \bigcup \bm{\mathcal{T}}_{l+}^{um},
\end{equation}
where
\begin{itemize}
    \item $\bm{\mathcal{T}}_{l+}^{mt}$: The set of matched and fused local tracks, see Line 1 to 12 in Algorithm \ref{table:track2track},
    \item $\bm{\mathcal{T}}_{l+}^{um}$: The set of unmatched local tracks after CI fusion update, see Line 13 to 23 in Algorithm \ref{table:track2track},
    \item $\bm{\mathcal{T}}_{l+}^{ud}$: The set of unmatched and updated local tracks that are not detected by the transmitting ITS-S, see Line 24 to 27 in Algorithm \ref{table:track2track},
\end{itemize}

The algorithm also creates and uses an interim variable $T_{l}^{nw}$, which denotes the new local track initiated by an unmatched remote track, see Line 15 to 16 in Algorithm \ref{table:track2track}. Also, one can choose to use a few post-processing strategies to effectively control the growth of tracks in $\bm{\mathcal{T}}_{l+}$:
\begin{enumerate}
    \item A pruning process can be introduced by removing tracks with weights lower than a threshold,
    \item A merging process can be used for combining close tracks based on some distance metrics, for instance, Mahalanobis distance,
    \item A maximum number of updated local tracks at each iteration can be set.
\end{enumerate}

\begin{algorithm}[!t]
\DontPrintSemicolon
\small{
\KwData{$\bm{\mathcal{T}}_{l}$, $\bm{\mathcal{T}}_{r}$}
\KwResult{$\bm{\mathcal{T}}_{l+}$}
$\bm{\mathcal{T}}_{l+}^{mt} \gets \emptyset$;
$\bm{\mathcal{T}}_{l}^{um} \gets \bm{\mathcal{T}}_{l}$;
$\bm{\mathcal{T}}_{r}^{um} \gets \bm{\mathcal{T}}_{r}$\;
\ForEach{$T_{r} \in \bm{\mathcal{T}}_{r}$}{
    $\bm{\mathcal{T}}_{+} \gets \emptyset$\;
    \ForEach{$T_{l} \in \bm{\mathcal{T}}_{l}$}{
        \If{$\begin{bmatrix}s_{r}&d_{r}\end{bmatrix} \in \textbf{R}$}{
            $T_{l+} \gets T_{l}$\;
            $\left\{\bar{\textbf{x}}_{l+}, \bm{\Sigma}_{l+}, w_{+}\right\} \gets CI\left(T_{l}, T_{r}\right)$\;
            $\bm{\mathcal{T}}_{+} \gets \bm{\mathcal{T}}_{+} \bigcup \left\{T_{l+}\right\}$\;
            $\bm{\mathcal{T}}_{l}^{um} \gets \bm{\mathcal{T}}_{l}^{um} \backslash \left\{T_{l}\right\}$\;
            $\bm{\mathcal{T}}_{r}^{um} \gets \bm{\mathcal{T}}_{r}^{um} \backslash \left\{T_{r}\right\}$\;
        }
    }
    Normalise $\forall w_{+} \in \bm{\mathcal{T}}_{+}$\;
    $\bm{\mathcal{T}}_{l+}^{mt} \gets \bm{\mathcal{T}}_{l+}^{mt} \bigcup \bm{\mathcal{T}}_{+}$\;
}
$\bm{\mathcal{T}}_{l+}^{um} \gets \emptyset$\;
\ForEach{$T_{r} \in \bm{\mathcal{T}}_{r}^{um}$}{
    $w^{nw} \gets w_{\gamma}$;
    $d_{l}^{nw} \gets \text{New Track ID}$;
    $\textbf{R}^{nw} \gets \begin{bmatrix}s_{r}&d_{r}\end{bmatrix}$\;
    $T_{l}^{nw} \gets \left\{\bar{\textbf{x}}_{r}, \bm{\Sigma}_{r}, w^{nw}, d_{l}^{nw}, \textbf{R}^{nw}\right\}$;
    $\bm{\mathcal{T}}_{+} \gets T_{l}^{nw}$\;
    \ForEach{$T_{l} \in \bm{\mathcal{T}}_{l}^{um}$}{
        \If{$s_{r} \not\in \textbf{R}$}{
            $T_{l+} \gets T_{l}$;
            $\textbf{R}_{+} \gets \textbf{R} \bigcup \left\{\begin{bmatrix}s_{r}&d_{r}\end{bmatrix}\right\}$\;
            $\left\{\bar{\textbf{x}}_{l+}, \bm{\Sigma}_{l+}, w_{+}\right\} \gets CI\left(T_{l}, T_{r}\right)$\;
            $\bm{\mathcal{T}}_{+} \gets \bm{\mathcal{T}}_{+} \bigcup \left\{T_{l+}\right\}$\;
        }
    }
    Normalise $\forall w_{+} \in \bm{\mathcal{T}}_{+}$\;
    $\bm{\mathcal{T}}_{l+}^{um} \gets \bm{\mathcal{T}}_{l+}^{um} \bigcup \bm{\mathcal{T}}_{+}$\;
}
$\bm{\mathcal{T}}_{l+}^{ud} \gets \emptyset$\;
\ForEach{$T_{l} \in \bm{\mathcal{T}}_{l}^{um}$}{
    $T_{l+} \gets T_{l}$;
    $w_{+} \gets \left(1-p_{D,r}\left(\bar{\textbf{x}}_{l}\right)\right) w$\;
    $\bm{\mathcal{T}}_{l+}^{ud} \gets \bm{\mathcal{T}}_{l+}^{ud} \bigcup \left\{T_{l+}\right\}$\;
}
$\bm{\mathcal{T}}_{l+} \gets \bm{\mathcal{T}}_{l+}^{ud} \bigcup \bm{\mathcal{T}}_{l+}^{mt} \bigcup \bm{\mathcal{T}}_{l+}^{um}$\;
}
 \caption{Track-To-Track Fusion}
 \label{table:track2track}
\end{algorithm}

As previously mentioned, the track-to-track fusion can be required between the host CV and multiple other ITS-Ss within the same fusion cycle. This can be done by performing $\bm{\mathcal{T}}_{l} \gets \bm{\mathcal{T}}_{l+}$ and running Algorithm \ref{table:track2track} for $\bm{\mathcal{T}}_{l}$ and $\bm{\mathcal{T}}_{r}$ from every other cooperative ITS-S, iteratively.

\section{Results}
\label{sec:results}

\subsection{Numerical Simulation}
\label{sec:num_sim}

\begin{figure}[!t]
	\centering
	\includegraphics[width=2.4in, trim={1.7in 2.8in 6.7in 2in}, clip]{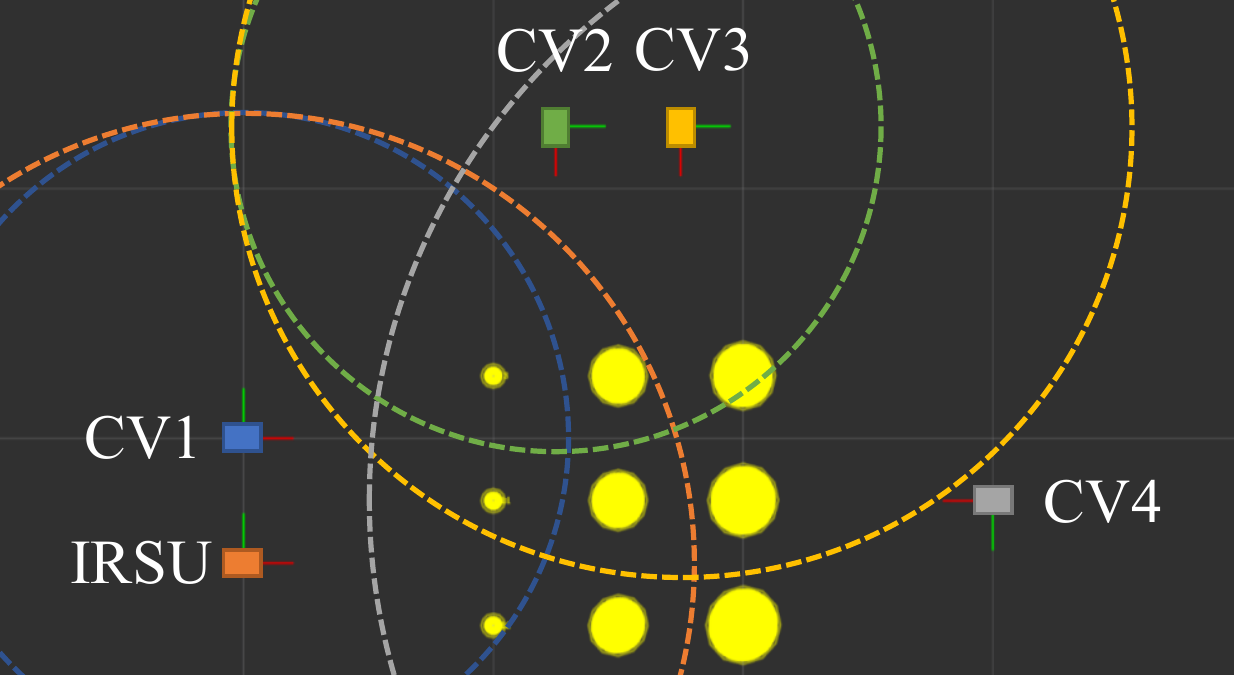}
	\caption{The setup of the numerical simulation and the qualitative tracking results for \textit{Configuration A}. The simulation contains 4 CVs and one IRSU. The 9 pedestrians are arranged as a $3\times 3$ grid, at an interval of 5 m. The dotted circles represent the sensing range of each ITS-S. Different sensing ranges are set for the ITS-Ss such that each pedestrian corresponds to one of 9 different combinations of perception coverage of the ITS-Ss. The setup and results are visualised in the local coordinate system of CV1. The 95\% confidence ellipse for the 2D position estimate of each tracked pedestrian as a result of V2X data fusion is shown in yellow.}
	\label{fig:num_sim_setup_result}
\end{figure}

The proposed data fusion framework for CP is first validated through numerical simulation. As illustrated in Figure \ref{fig:num_sim_setup_result}, the simulation is set up with 5 ITS-Ss, including 4 CVs and one IRSU, sensing a group of static pedestrians within their local frames with some level of uncertainty in the perception process. The positions of the pedestrians and the perception range of each ITS-S are arranged deliberately so that different subsets of the pedestrians are detectable by different ITS-Ss. This also means a unique combination of ITS-Ss perceives every pedestrian. There is bi-directional \textit{vehicle-to-vehicle} (V2V) communication established between each pair of CVs and \textit{vehicle-to-infrastructure} (V2I) communication from the IRSU to each CV. All the ITS-Ss are assumed to contain pose uncertainty.

Within each ITS-S, the perceived objects information and egocentric pose estimate are encoded into CPMs in the form of binary payloads and published at 10 Hz. Within each CV, an instance of road user tracker is running, and the tracked pedestrian estimates are encoded and included in the CPMs. Each CV also decodes the CPMs received from other ITS-Ss and transforms the perceived objects into its local coordinate system before being fused in its road user tracker. Also, there are two configurations considered for the IRSU for comparison. In \textit{Configuration A}, the IRSU is using the tracked pedestrian estimates as the perceived objects information in the CPMs just like the case in the CVs, while in \textit{Configuration B}, the IRSU is publishing the independent pedestrian sensor detections instead.

The ITS-Ss perceive each pedestrian with a standard deviation of $0.2 m$ in the 2D position measurement. For every CV, the standard deviations for the position and heading estimates in the localisation are set to $0.25 m$ and $0.5^{\circ}$, respectively. Note that for the IRSU, the localisation noise is assumed small yet non-zero. The standard deviation for the position estimate of the IRSU is set to $0.005 m$, which is roughly the minimum value supported in CPMs for representing the uncertainty of position estimates, and its heading estimate standard deviation is set to a close-to-zero value $\epsilon$ for preventing numerical errors in the coordinate transformation. Each pedestrian also employs a simple kinematic model with independent linear velocity noise with a standard deviation of $1.0 m/s$ in both $x$ and $y$. The simulation is run for 10 s before the estimation results are obtained, ensuring the data fusion has reached a steady-state among all ITS-Ss.

Figure \ref{fig:num_sim_setup_result} and Table \ref{tab:sim_result_config_a} reveal the qualitative and quantitative estimation results, respectively, of \textit{Configuration A} in the local frame of CV1. It is demonstrated that the proposed data fusion framework manages to correctly and safely fuse the remote object tracks, with unknown cross-correlation, from the mixture of 4 CVs and one IRSU. It also clearly shows how the data fusion reduces the estimate uncertainty by fusing perception data from more ITS-Ss. The top-left pedestrian in Figure \ref{fig:num_sim_setup_result} has the most accurate position estimate result, as it is observed by all five ITS-Ss, while the bottom-right one is only observed by one CV and has the largest estimate uncertainty of all.

Table \ref{tab:sim_result_config_b} depicts the results of \textit{Configuration B}, where the IRSU is sharing its local pedestrian detections instead of tracking results. While the conclusion drawn from \textit{Configuration A} remains the same in \textit{Configuration B}, it can also be seen from the results that this configuration dramatically reduces the estimate uncertainty level for the pedestrians in the middle column, which are detectable by the IRSU but stay out of the perception range of CV1. The reduction in estimate uncertainty is also observed for every other pedestrian in this configuration as a result of the fusion process, compared with that in \textit{Configuration A}.
However, this finding does not suggest that \textit{Configuration B} should also be adopted in the CVs as their information sharing mode, because 1) an object track contains extra properties gained through tracking, such as velocity and heading, which can improve the quality of estimates, 2) it comprises data fusion outcome up to the present time and thus sharing them also saves V2V communication load.

\begin{table}[!t]
\caption{Standard deviations in position estimate for \textit{Configuration A}}
\centering
\scalebox{0.9}{
\begin{tabular}{cccc}
\toprule
& \multicolumn{3}{c}{Standard Deviation in Position $x$/$y$ (m)} \\
& Column 1 & Column 2 & Column 3 \\
\midrule
Row 1 & 0.11554/0.11558 & 0.42214/0.44471 & 0.47780/0.50964\\
Row 2 & 0.11560/0.11564 & 0.42265/0.44575 & 0.52531/0.55300\\
Row 3 & 0.11568/0.11569 & 0.42861/0.44607 & 0.56028/0.58923\\
\bottomrule
\end{tabular}}
\label{tab:sim_result_config_a}
\end{table}

\begin{table}[!t]
\caption{Standard deviations in position estimate for \textit{Configuration B}}
\centering
\scalebox{0.9}{
\begin{tabular}{cccc}
\toprule
& \multicolumn{3}{c}{Standard Deviation in Position $x$/$y$ (m)} \\
& Column 1 & Column 2 & Column 3 \\
\midrule
Row 1 & 0.10622/0.10654 & 0.16934/0.17598 & 0.46924/0.50121\\
Row 2 & 0.10624/0.10655 & 0.16949/0.17607 & 0.51678/0.54458\\
Row 3 & 0.10642/0.10655 & 0.17120/0.17610 & 0.55911/0.58759\\
\bottomrule
\end{tabular}}
\label{tab:sim_result_config_b}
\end{table}

\subsection{CARLA Simulation}

\begin{figure}[!t]
	\centering
	\includegraphics[width=2.7in, trim={0in 4.45in 6.7in 0in}, clip]{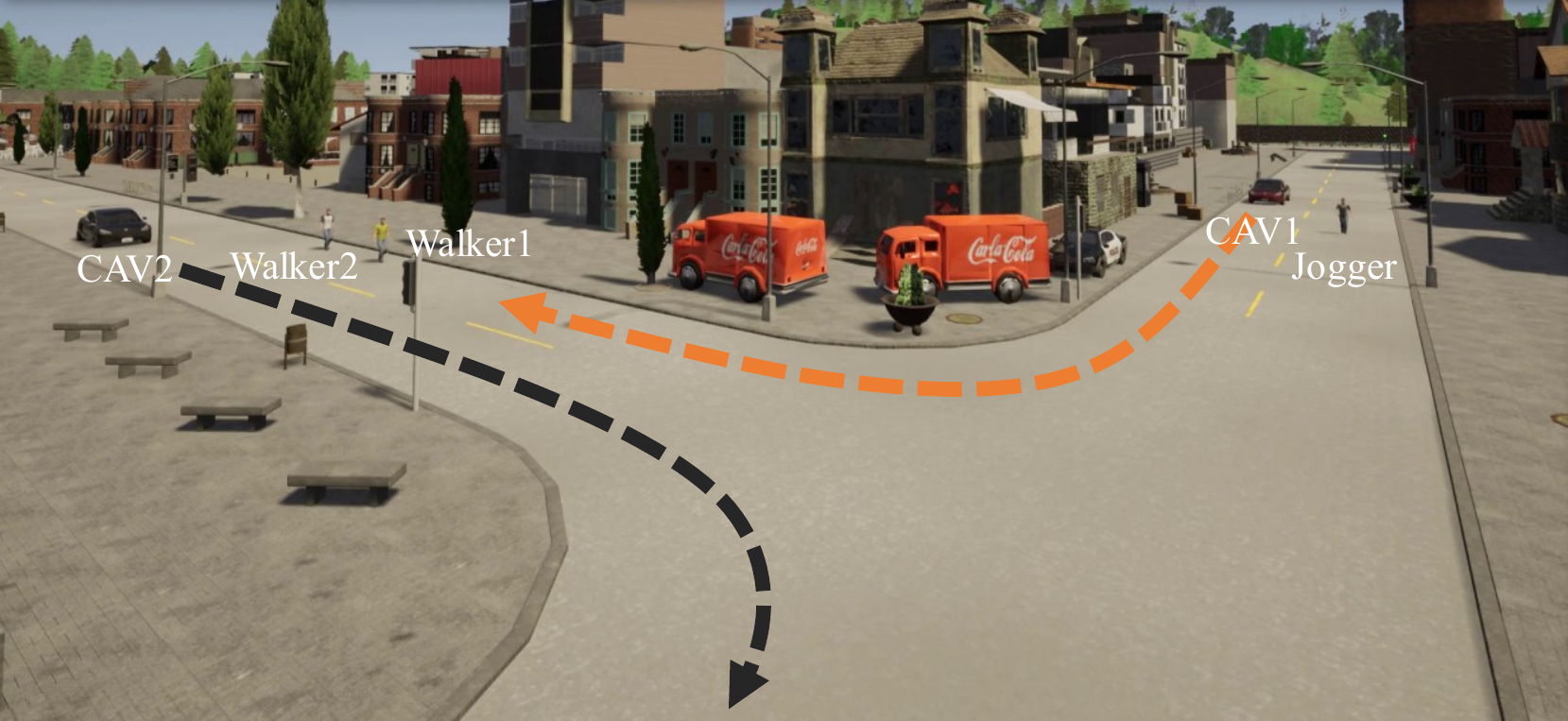}
	\caption{Simulation setup in the CARLA. Two CAVs were moving towards an intersection before turning right, as the dashed lines indicate. There were two walkers and a jogger within the respective detection range of the CAVs.}
	\label{fig:carla_setup}
\end{figure}

The simulation was set up in the \textit{Town01} map of CARLA simulator with two moving CAVs and three walking/running pedestrians, as labelled in Figure \ref{fig:carla_setup}. The CAVs approached an intersection from different directions before making a right turn. A few other cars were parked next to the intersection to create a visual obstruction for each CAV. It is assumed that there was no traffic control setup at this intersection. It is also configured in the simulator that the self-localisation of the CAVs was not perfect, and an \textit{unscented Kalman filter} (UKF) is employed to fuse the noise-corrupted measurements from GNSS, IMU, and wheel encoder. As a result, the standard deviation in position estimate ranges from 0.11 to 0.15 m, and that for the heading fluctuates between 0.025 and 0.035 rad over time. Besides, both CAVs were equipped with a front-facing camera and a 16-beam lidar for road user detection and tracking, and there were ETSI CPMs exchanged between them at 10 Hz. CAV1 was navigating autonomously on the road using the navigation stack presented in \cite{paper:NarulaWorrall2020}, while CAV2 followed a predetermined path. In the simulation, CAV1 was moving up to 18 km/h in autonomous navigation while CAV2 was programmed to move at 10 km/h. The two walkers moved at 2 m/s, while the jogger moved at 2.8 m/s, all towards the intersection. A constant velocity model is employed as their kinematic model in the trackers.

In the simulation, each CAV fused its local sensory information and the perception data transmitted by the other CAV through V2V communication using the proposed data fusion framework. There is an instance of the road user tracker running within each CAV for tracking the states of the pedestrians within its local frame, including their position, heading, and speed, based on the pedestrian detections from the fusion of lidar and camera, and remote object tracks it receives from the other CAV.

\begin{figure}[!t]
	\centering
	\subfigure[]{ 
		\label{fig:carla_tracking_result:a} 
		\includegraphics[width=2.7in]{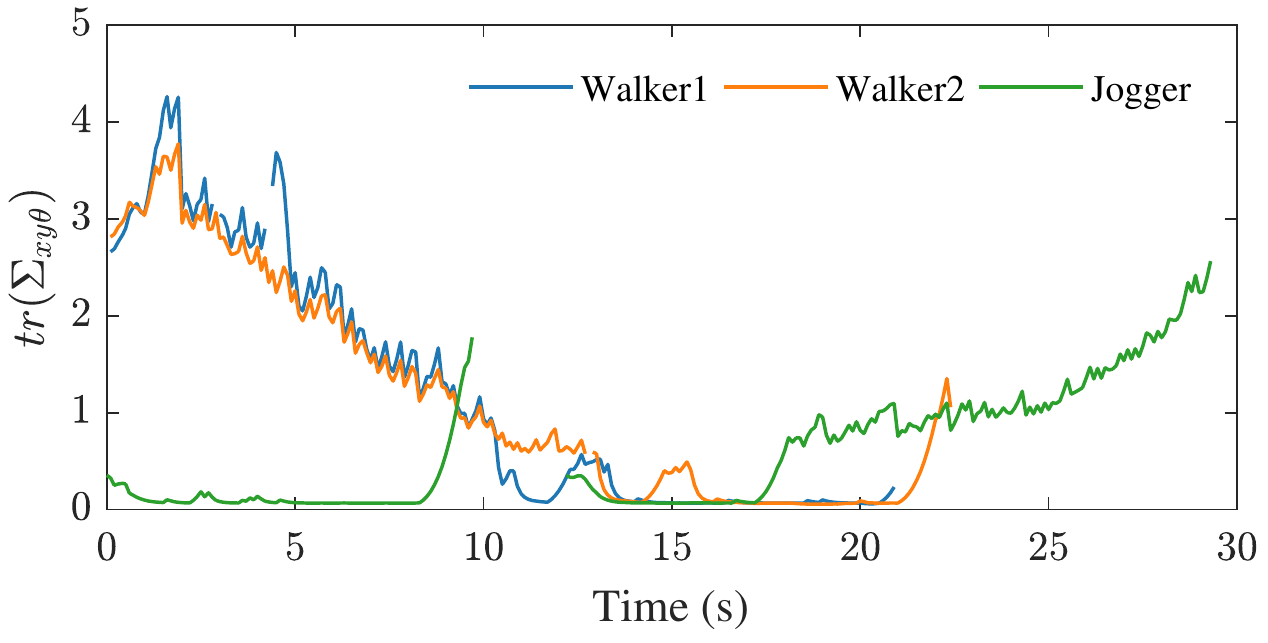}}
	\subfigure[]{ 
		\label{fig:carla_tracking_result:b} 
		\includegraphics[width=2.7in]{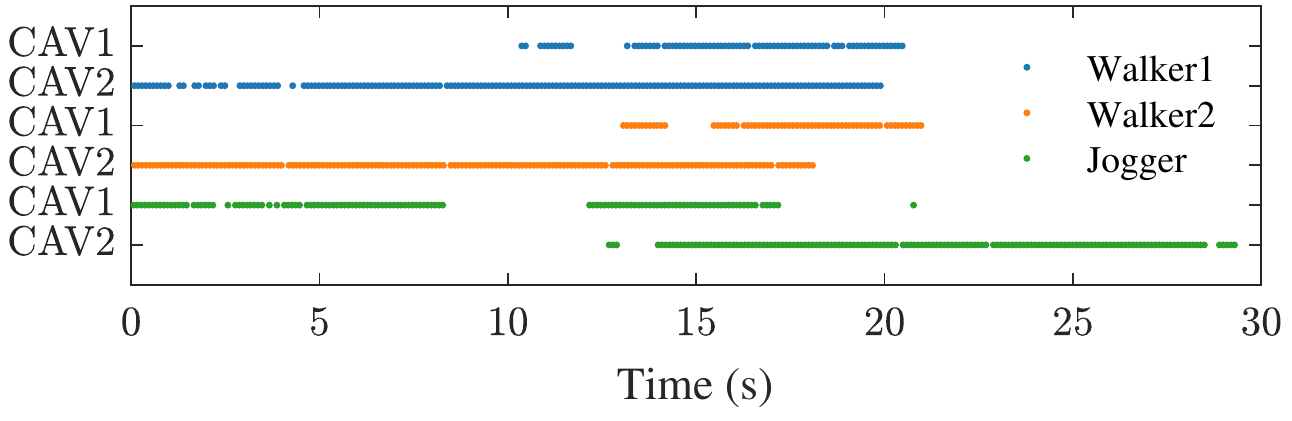}}
	\caption{V2X data fusion results in the simulation from the perspective of CAV1. (a) presents how the estimate uncertainty of the three pedestrians evolves over time, and (b) marks the time points when the local sensor detections of the pedestrians were available in CAV1.}
	\label{fig:carla_tracking_result} 
\end{figure}

Figure \ref{fig:carla_tracking_result} presents the V2X data fusion results over time in the simulation. The uncertainty of each tracked pedestrian is evaluated quantitatively using $tr\left(\Sigma_{xy\theta}\right)$, which is the trace of the covariance matrix of the $x$ and $y$ positions and heading $\theta$ components in the state estimate of the pedestrian. Depending on the evolvement of perception situation of the pedestrians, the simulation timeline can be roughly divided into three parts as follows.

\begin{figure}[t!]
	\centering
	\begin{minipage}[c]{1.11in}
        \centering
        \subfigure[]{
            \label{fig:carla_tracking_result_time1:a} 
            \includegraphics[width=1.1in]{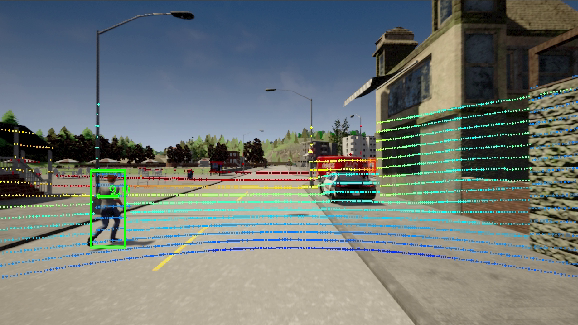}}
        \subfigure[]{
            \label{fig:carla_tracking_result_time1:b} 
            \includegraphics[width=1.1in]{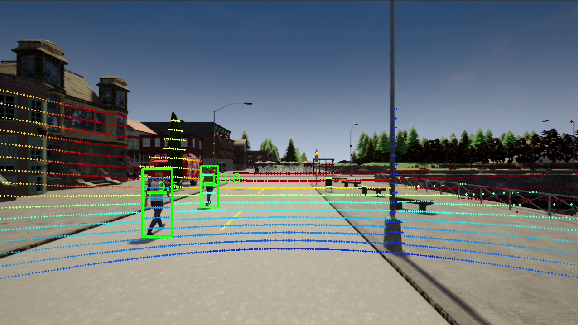}}
    \end{minipage}
    \begin{minipage}[c]{1.86in}
        \centering
        \subfigure[]{
            \label{fig:carla_tracking_result_time1:c} 
            \includegraphics[width=1.85in, trim={3.5in 0in 1in 1.5in}, clip]{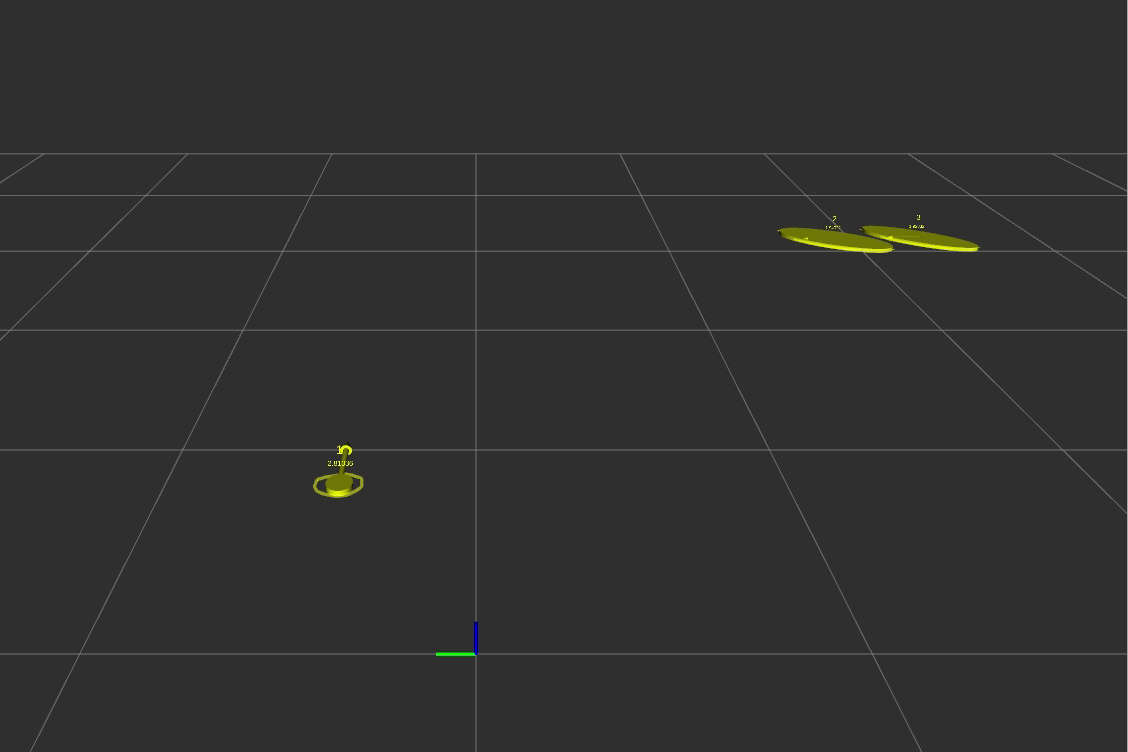}}
    \end{minipage}
	\caption{Sensory data processing and pedestrian tracking in CAV1 at simulation time 6.90 s. (a) and (b) illustrate the sensor fusion of onboard camera and lidar and detection of road users in CAV1 and CAV2, respectively. As shown in (a), CAV1 was able to observe the jogger, but could not detect the two walkers behind the building. In (b), CAV2 were perceiving the two walkers while not able to detect the jogger. In (c), CAV1 was able to track all three pedestrians using the proposed data fusion method based on local sensory detection and tracking information from CAV2 through V2V communication. The tracked pedestrians are visualised with 95\% confidence ellipses and arrows indicating their moving direction. Likewise, although not shown here, CAV2 were tracking the same three pedestrians within its local frame by fusing local and remote perception information.}
	\label{fig:carla_tracking_result_time1}
\end{figure}

\begin{figure}[!t]
	\centering
	\subfigure[]{
        \label{fig:carla_tracking_result_time2:a} 
        \includegraphics[width=1.3in]{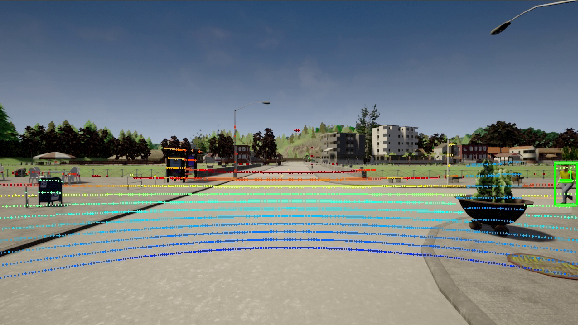}}
    \subfigure[]{
        \label{fig:carla_tracking_result_time2:b} 
        \includegraphics[width=1.3in]{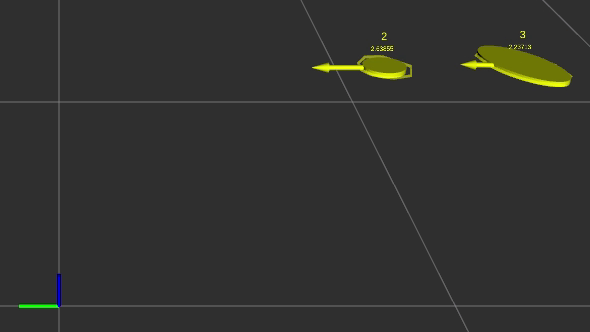}}
    \subfigure[]{
        \label{fig:carla_tracking_result_time2:c} 
        \includegraphics[width=1.3in]{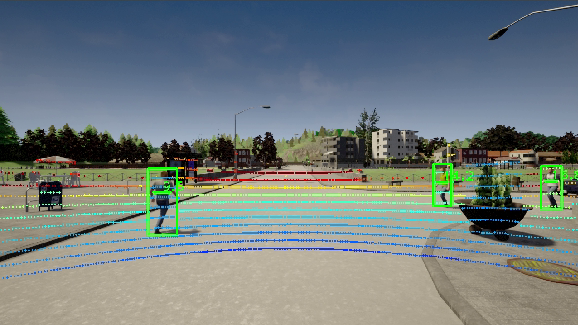}}
    \subfigure[]{
        \label{fig:carla_tracking_result_time2:d} 
        \includegraphics[width=1.3in]{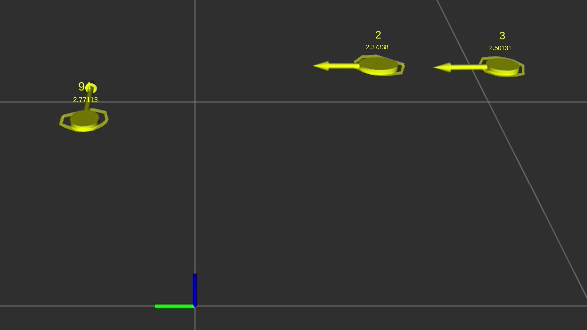}}
	\caption{Pedestrian detection and tracking in CAV1 at time 10.99 s in (a), (b), and 13.59 s in (c), (d). In (a), with the detection of Walker1 using its own sensors, CAV1 managed to reduce the tracking uncertainty by fusing the local detection into tracking, as shown in (b). In (c), CAV1 could detect both walkers using the onboard perception sensors. As a result, as shown in (d), the tracking uncertainty of Walker2 dropped significantly too.}
	\label{fig:carla_tracking_result_time2}
\end{figure}

\begin{itemize}
    \item From the start to simulation time 10.37 s, the CAV1 could not detect the two walkers using its local sensors, yet it managed to continuously track them in its local coordinate systems using the perception information received from CAV2. Figure \ref{fig:carla_tracking_result_time1} presents the details of local sensory data and the tracking results in CAV1 at time 6.90 s. It is also shown in Figure \ref{fig:carla_tracking_result:a} that during this time period, the $tr\left(\Sigma_{xy\theta}\right)$ of each tracked walker reduces steadily as the walkers and CAVs were moving closer to the intersection.
    \item From time 10.37 to 20.97 s, the CAV1 started to be able to detect the walkers directly when they walked out from the visual occlusion. Figure \ref{fig:carla_tracking_result_time2} illustrates the significant drop of the tracking uncertainty for Walker1 and Walker2 at time 10.99 s and 13.59 s, respectively, in CAV1 when the corresponding local sensor detections were available for the pedestrian tracker, as Figure \ref{fig:carla_tracking_result:b} also shows.
    \item From time 17.17 s to the end, the CAV1 lost local detection of the jogger due to the pedestrian moving out of the sensor range, yet it still maintained the tracking of the jogger thanks to the tracking information received from CAV2. Nevertheless, as Figure \ref{fig:carla_tracking_result:a} depicts, the tracking uncertainty for the jogger grew slowly in CAV1, due to the absence of local sensor detection for the pedestrian.
\end{itemize}

It is important to notice in the simulation results that despite sharing of the common tracking information between the two CAVs multiple times, the CI algorithm adopted for fusing local and remote object tracks, which are estimates with unknown cross-correlation, within the pedestrian tracker effectively avoids the inconsistent estimation issue.

\begin{figure}[!t]
	\centering
	\subfigure[]{
        \label{fig:carla_planning:a} 
        \includegraphics[width=0.8in]{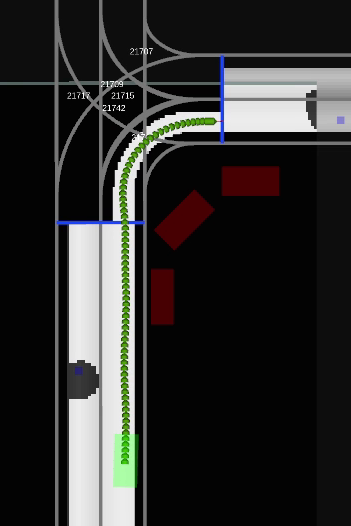}}
    \subfigure[]{
        \label{fig:carla_planning:b} 
        \includegraphics[width=0.8in]{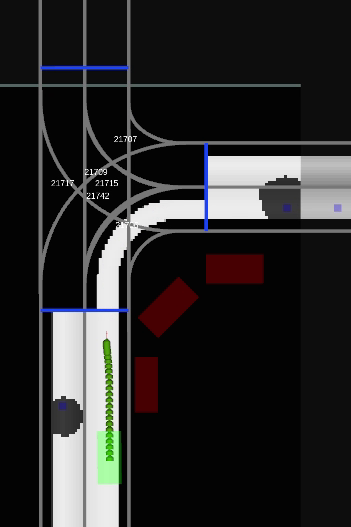}}
    \subfigure[]{
        \label{fig:carla_planning:c} 
        \includegraphics[width=0.8in]{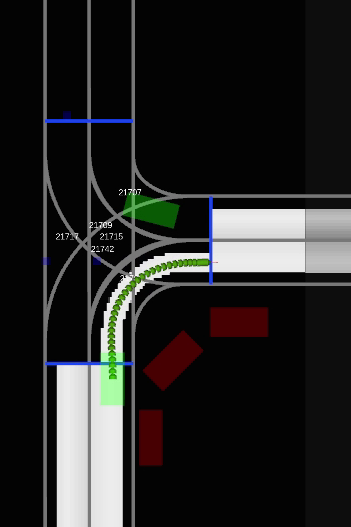}}
    \subfigure[]{
        \label{fig:carla_planning:d} 
        \includegraphics[width=2.8in]{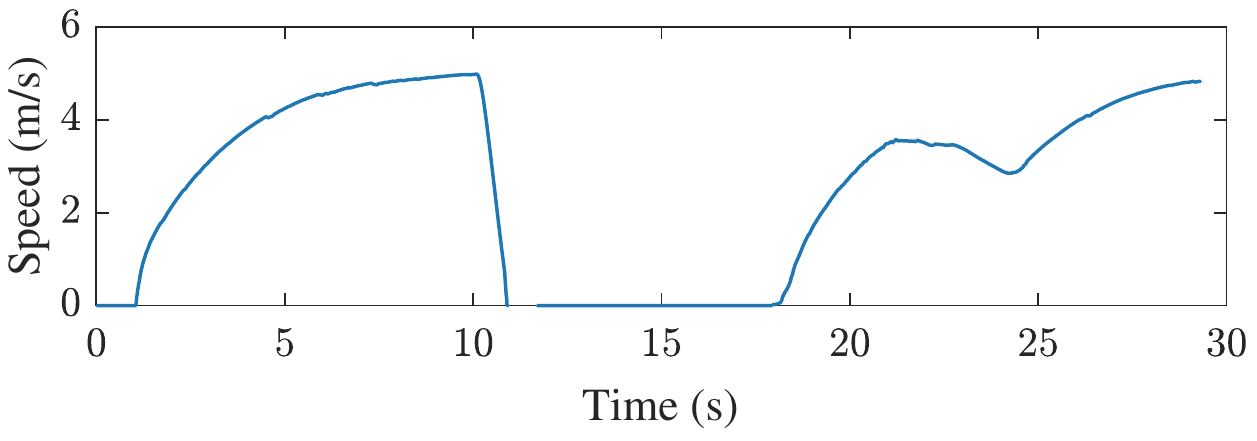}}
	\caption{The planned path for CAV1 at different simulation times. a) at time 6.90 s, a path (the green dotted curve) was planned within CAV1 for turning right at the intersection. Later at time 8.64 s, CAV1 learnt that the two walkers were about to cross the intersection based on the tracking information received from CAV2. In this case, CAV1 decided to give way to the walkers as it could no more find a feasible path to make the right turn at the intersection, as shown in (b). Up to this point, CAV1 still could not observe the pedestrians using its local sensors. It later stopped in front of the intersection, as (d) shows. Later at time 19.78 s, as the area for turning right became clear at the intersection, CAV1 planned a path and started turning in (c). The speed of CAV1 over time is presented in (d).}
	\label{fig:carla_planning}
\end{figure}

The simulation also shows how CP can benefit the path planning and decision making in the CAV and thus improve safety. As illustrated in Figure \ref{fig:carla_planning}, there was a feasible path planned within CAV1 for turning right at the intersection until time 8.64 s, when the two walkers were moving closer to the intersection to a point that the path planner could not find a feasible path to make the turn. In this case, CAV1 decided to give way to the crossing walkers and prepared to make a full stop later in front of the intersection, as depicted in Figure \ref{fig:carla_planning:b}. Please note that up to this point, CAV1 could not observe the two walking pedestrians directly using its local sensors due to the visual occlusion, yet CAV1 safely responded to the crossing behaviour of the pedestrians solely based on V2X perception information received from CAV2.

\subsection{Lab Experiment}

\begin{figure}[!t]
	\centering
	\includegraphics[width=2.9in, trim={2.6in 6.1in 4.3in 2.3in}, clip]{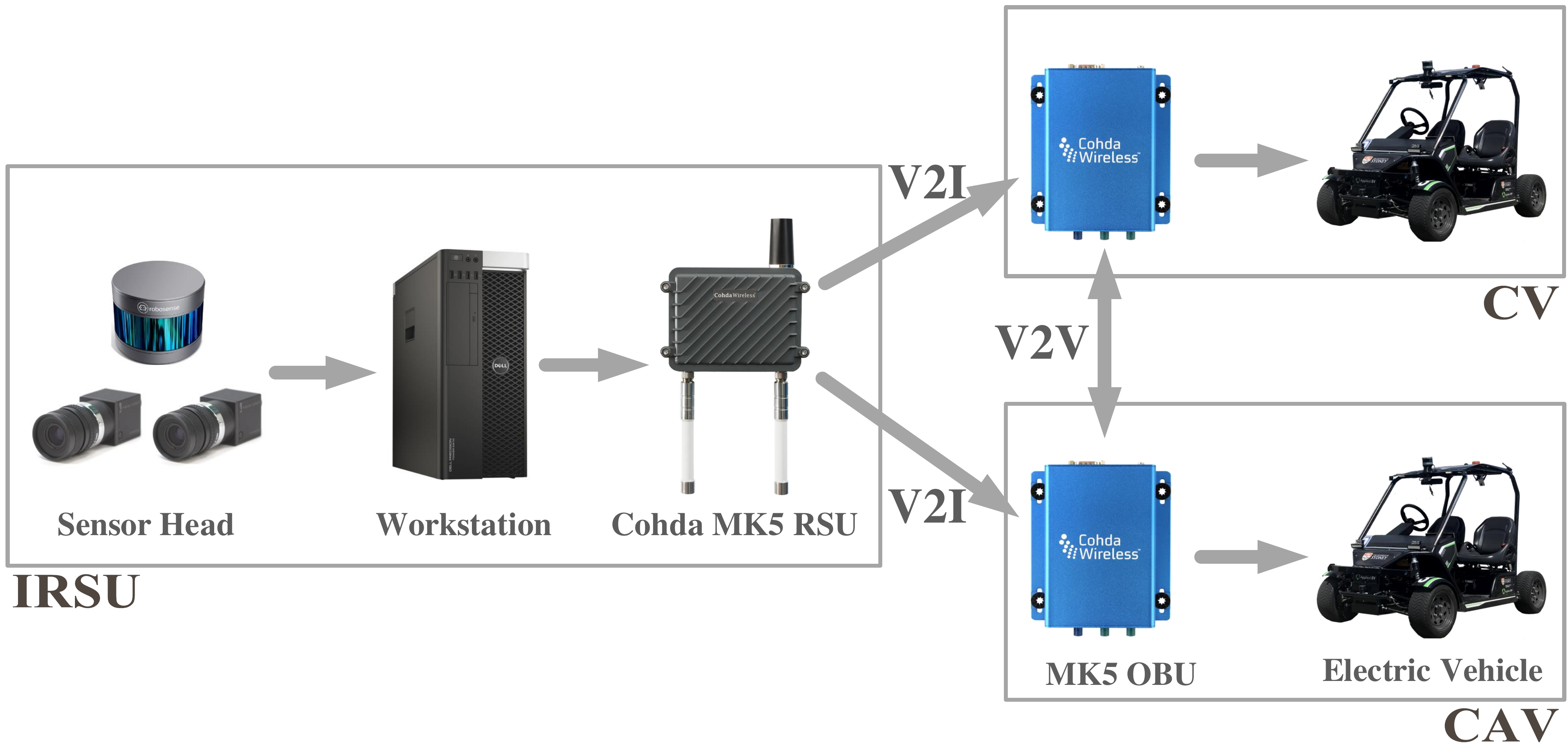}
	\caption{V2X communication topology in the lab experiment. The vehicles and the IRSU have V2X communication capability through Cohda Wireless MK5 OBU and RSU, respectively. The CV and CAV were exchanging CPMs through V2V communication, while the IRSU was transmitting CPMs to both vehicles via V2I communication.}
	\label{fig:lab_comm_setup}
\end{figure}

The section presents the real-world experiment results of a CAV using the fused V2X perception information from a CV and an IRSU and considered it in its automated operation in a lab environment. The communication topology of the participant ITS-Ss in the experiment is illustrated in Figure \ref{fig:lab_comm_setup}. As Figure \ref{fig:lab_comm_setup} reveals, the two vehicles were exchanging V2X data fusion results in the form of ETSI CPMs through V2V communication, and in the meantime, receiving CPMs from the IRSU through V2I communication. The IRSU was setup in the experiment to populate CPMs with object detection information instead of object tracks, i.e., following \textit{Configuration B} in Section \ref{sec:num_sim}.

Both vehicle platforms are equipped with a suite of local perception sensors, and the CAV also has the navigation stack in \cite{paper:NarulaWorrall2020} running onboard to implement full autonomy. Nevertheless, the CAV did not use the internal perception capabilities in its automated operation in the experiment so as to highlight the safety benefits of using the CP service, in particular, under the circumstances of persistent visual occlusion or perception sensor malfunction. The experiment site has been surveyed with a \textit{Lanelet2} map built. Both vehicles were using RTK to localise themselves.

\begin{figure}[!t]
	\centering
	\subfigure[]{ 
		\label{fig:lab_setup:a} 
		\includegraphics[width=2.7in, trim={0in 5.1in 7.4in 0in}, clip]{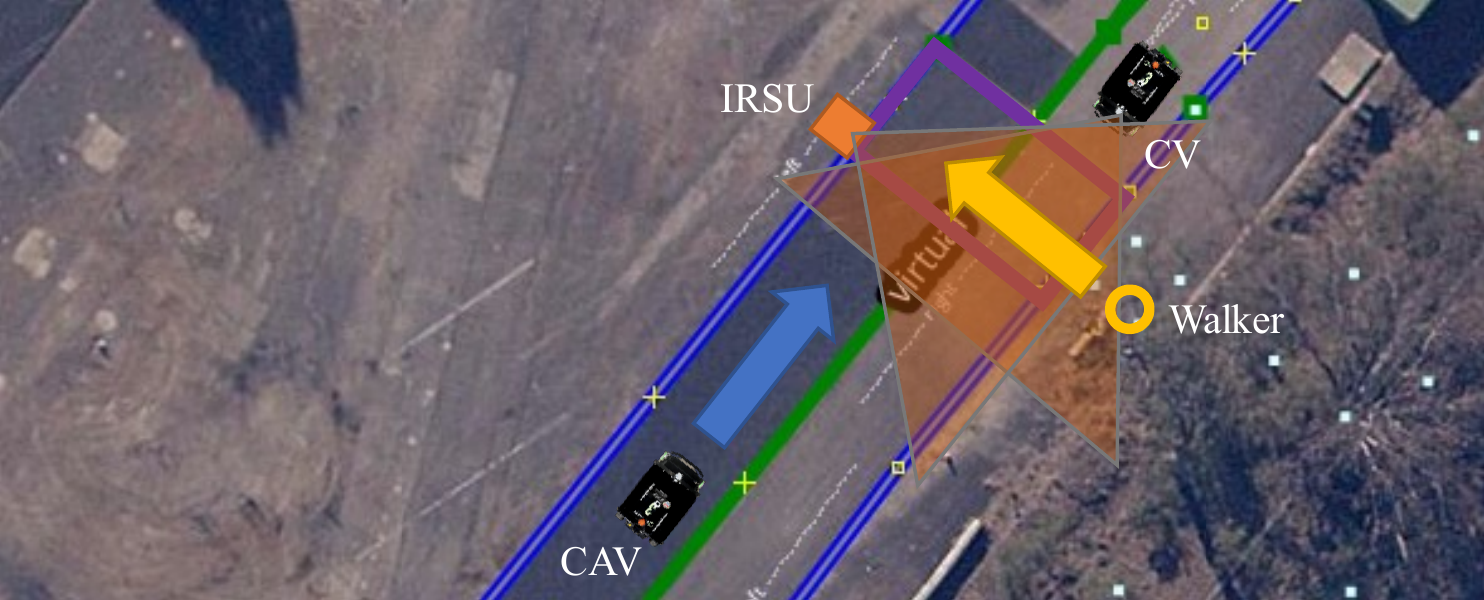}}
	\subfigure[]{ 
		\label{fig:lab_setup:b} 
		\includegraphics[width=2.7in, trim={0in 5.1in 7.4in 0in}, clip]{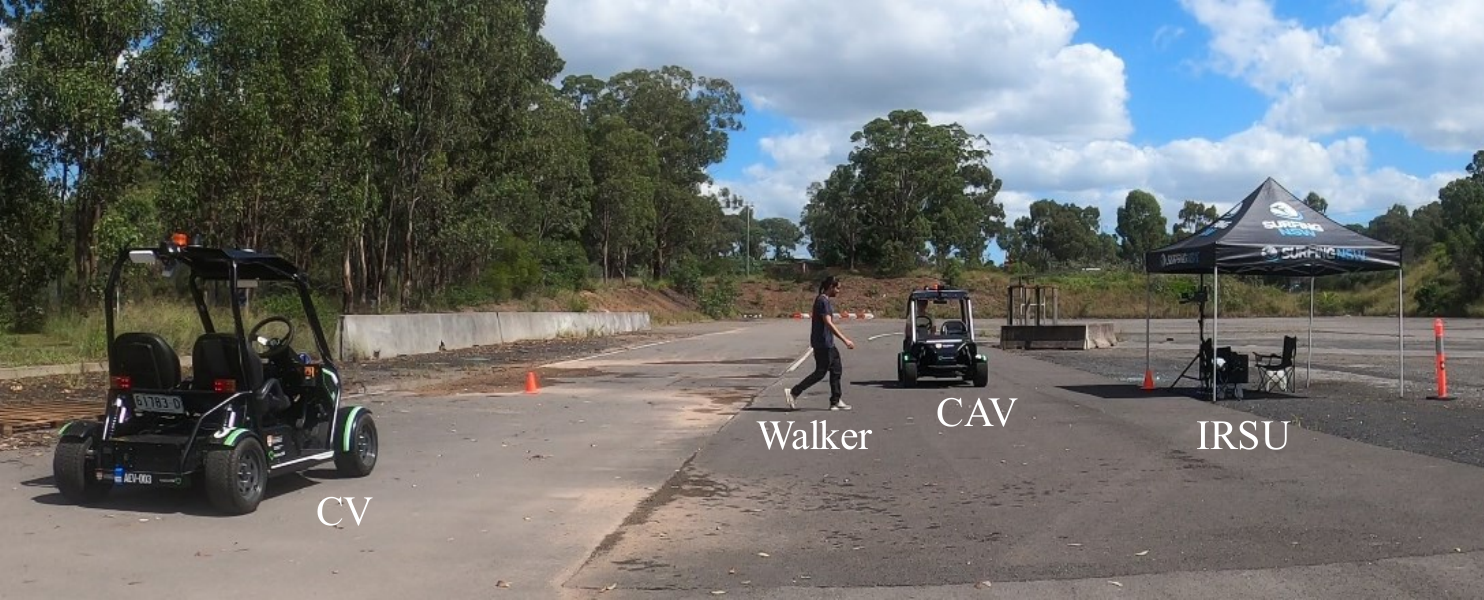}}
	\caption{Experiment setup in the lab environment. (a) shows the satellite map of the lab annotated with the road boundaries, lane information, and a designated pedestrian crossing area (in purple). The map also shows the CV and the IRSU near the pedestrian crossing, and the CAV autonomously moving northeast towards the crossing.}
	\label{fig:lab_setup}
\end{figure}

\begin{figure}[!t]
	\centering
	\includegraphics[width=2.7in, trim={0in 0in 0in 0in}, clip]{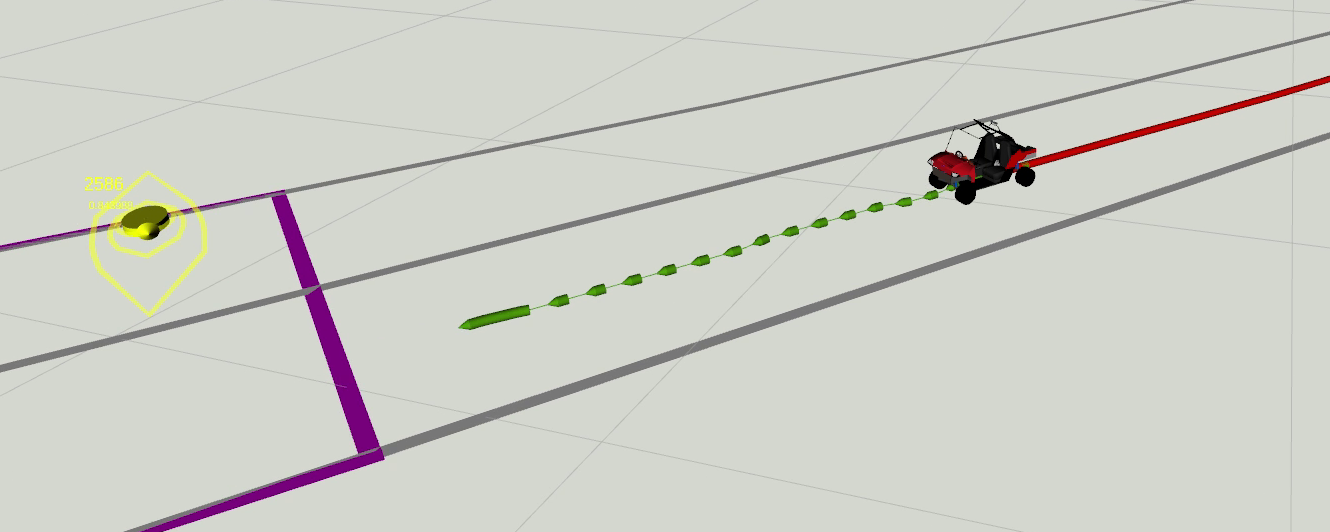}
	\caption{The CAV approaching the pedestrian crossing while it became aware of the crossing pedestrian. This information is provided through the fusion of perception information from the CV and the IRSU.}
	\label{fig:lab_result_cav}
\end{figure}

\begin{figure}[!t]
	\centering
	\subfigure[]{ 
		\label{fig:lab_result_fusion:a} 
		\includegraphics[width=1.4in, trim={0in 0.1in 0in 0in}, clip]{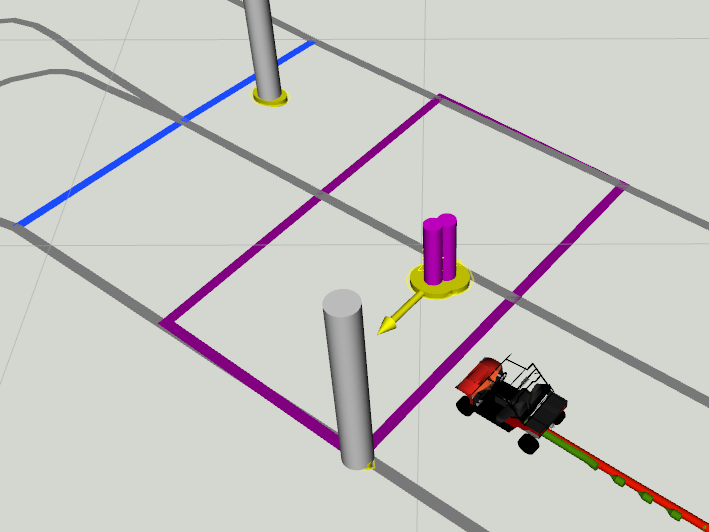}}
	\subfigure[]{ 
		\label{fig:lab_result_fusion:b} 
		\includegraphics[width=1.4in, trim={0in 0.1in 0in 0in}, clip]{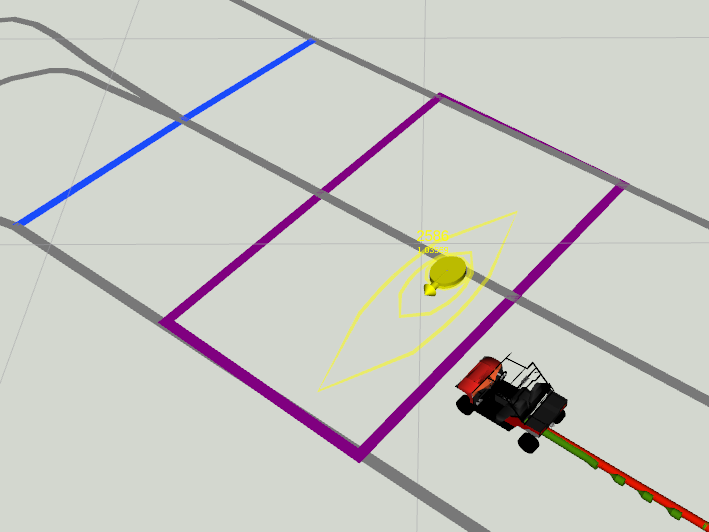}}
	\caption{Comparison of before and after fusion in the CAV when stopping in front of the crossing for the walker. The perception information before fusion is shown in (a), where there are two purple pillars corresponding to the track and the detection of the same target received from the CV and the IRSU, respectively. The tall grey pillars represent the locations of the CV and the IRSU. (b) shows the fused perception information in the CAV, presented as a single Gaussian with lower uncertainty (the inner yellow disc). The two outer yellow hollow ellipses represent future motion prediction of the walker.}
	\label{fig:lab_result_fusion}
\end{figure}

The experiment at the lab was setup as shown in Figure \ref{fig:lab_setup}. It can be seen that there was an overlap of perception areas of the CV and the IRSU. As depicted in Figure \ref{fig:lab_result_cav}, with the perception information from the CV and IRSU received, the CAV performed data fusion and managed to track the walker crossing the road. Figure \ref{fig:lab_result_fusion} provides an example comparing before and after the data fusion, and illustrating the reduction in estimation uncertainty after the fusion. The automated operation of the CAV in the experiment was as expected---it slowed down and gave way to the crossing walker relying on the fused perception information from the IRSU and the CV, it then continued to move after the crossing was detected to be clear.

\section{Conclusions}
\label{sec:conlusions}

The paper presents a novel V2X data fusion framework for CP that can perform a consistent fusion of local perception sensory data and perceived objects information received from other ITS-Ss through V2X communication. The proposed approach is validated through simulations and a real-world experiment. It is highlighted in the results that a CAV could extend its perception range and increase perception quality using the proposed approach at an intersection in the CARLA simulator, and a pedestrian crossing in the real-world experiment. The results also showcase the safety benefits of CP for CAV operations, and demonstrate how a CAV can consider the fused V2X perception information in its path planning and decision making when interacting with VRU.

\bibliography{main}

\begin{thebibliography}{10}
\providecommand{\url}[1]{#1}
\csname url@samestyle\endcsname
\providecommand{\newblock}{\relax}
\providecommand{\bibinfo}[2]{#2}
\providecommand{\BIBentrySTDinterwordspacing}{\spaceskip=0pt\relax}
\providecommand{\BIBentryALTinterwordstretchfactor}{4}
\providecommand{\BIBentryALTinterwordspacing}{\spaceskip=\fontdimen2\font plus
\BIBentryALTinterwordstretchfactor\fontdimen3\font minus
  \fontdimen4\font\relax}
\providecommand{\BIBforeignlanguage}[2]{{%
\expandafter\ifx\csname l@#1\endcsname\relax
\typeout{** WARNING: IEEEtran.bst: No hyphenation pattern has been}%
\typeout{** loaded for the language `#1'. Using the pattern for}%
\typeout{** the default language instead.}%
\else
\language=\csname l@#1\endcsname
\fi
#2}}
\providecommand{\BIBdecl}{\relax}
\BIBdecl

\bibitem{paper:RauchKlanner2012}
A.~Rauch, F.~Klanner, R.~Rasshofer, and K.~Dietmayer, ``Car2x-based perception
  in a high-level fusion architecture for cooperative perception systems,'' in
  \emph{Proc. the IEEE Intelligent Vehicles Symposium (IV)}, Alcal\'a de
  Henares, Spain, Jun. 2012, pp. 270--275.

\bibitem{web:KoFAS}
``Ko-{FAS},'' in \emph{Available online: http://www.ko-fas.de (accessed on 12
  March 2022)}.

\bibitem{paper:RauchMaier2013}
A.~Rauch, S.~Maier, F.~Klanner, and K.~Dietmayer, ``Inter-vehicle object
  association for cooperative perception systems,'' in \emph{Proc. the IEEE
  Intelligent Transportation Systems Conference (ITSC)}, The Hague, The
  Netherlands, Oct. 2013, pp. 893--898.

\bibitem{paper:MouawadMannoni2021_2}
N.~Mouawad and V.~Mannoni, ``Collective perception messages: New low complexity
  fusion and v2x connectivity analysis,'' in \emph{Proc. the IEEE 94th
  Vehicular Technology Conference (VTC2021-Fall)}, Norman, OK, USA, Sep. 2021,
  pp. 1--5.

\bibitem{paper:GuntherMennenga2016}
H.-J. G\"unther, B.~Mennenga, O.~Trauer, R.~Riebl, and L.~Wolf, ``Realizing
  collective perception in a vehicle,'' in \emph{Proc. the IEEE Vehicular
  Networking Conference (VNC)}, Columbus, OH, USA, Dec. 2016.

\bibitem{paper:MouawadMannoni2021_1}
N.~Mouawad, V.~Mannoni, B.~Denis, and A.~P. da~Silva, ``Impact of lte-v2x
  connectivity on global occupancy maps in a cooperative collision avoidance
  (coca) system,'' in \emph{Proc. the IEEE 93rd Vehicular Technology Conference
  (VTC2021-Spring)}, Helsinki, Finland, Apr. 2021, pp. 1--5.

\bibitem{paper:GodoyJimenez}
\BIBentryALTinterwordspacing
J.~Godoy, V.~Jim\'enez, A.~Artu\H{n}edo, and J.~Villagra, ``A grid-based
  framework for collective perception in autonomous vehicles,'' \emph{Sensors},
  vol.~21, no.~3, 2021. [Online]. Available:
  \url{https://www.mdpi.com/1424-8220/21/3/744}
\BIBentrySTDinterwordspacing

\bibitem{paper:JulierUhlmann1997}
S.~Julier and J.~K. Uhlmann, ``A non-divergent estimation algorithm in the
  presence of unknown correlations,'' in \emph{Proceedings of the American
  Control Conference}, Jun. 1997, pp. 2369--2373.

\bibitem{paper:AlligWanielik2019ITSC}
C.~Allig and G.~Wanielik, ``Dynamic dissemination method for collective
  perception,'' in \emph{Proc. the IEEE Intelligent Transportation Systems
  Conference (ITSC)}, Auckland, New Zealand, Oct. 2019, pp. 3756--3762.

\bibitem{paper:ShanNarula2021}
\BIBentryALTinterwordspacing
M.~Shan, K.~Narula, Y.~F. Wong, S.~Worrall, M.~Khan, P.~Alexander, and
  E.~Nebot, ``Demonstrations of cooperative perception: Safety and robustness
  in connected and automated vehicle operations,'' \emph{Sensors}, vol.~21,
  no.~1, 2021. [Online]. Available:
  \url{https://www.mdpi.com/1424-8220/21/1/200}
\BIBentrySTDinterwordspacing

\bibitem{paper:VoMa2006}
B.-T. Vo and W.-K. Ma, ``The {G}aussian mixture probability hypothesis density
  filter,'' \emph{IEEE Transactions on Signal Processing}, vol.~54, no.~11, p.
  4091–4104, Nov. 2006.

\bibitem{paper:ArambelRago2001}
P.~O. Arambel, C.~Rago, and R.~K. Mehra, ``Covariance intersection algorithm
  for distributed spacecraft state estimation,'' in \emph{Proceedings of the
  American Control Conference}, Arlington, VA, USA, Jun. 2001, pp. 4398--4403.

\bibitem{paper:NarulaWorrall2020}
K.~Narula, S.~Worrall, and E.~Nebot, ``Two-level hierarchical planning in a
  known semi-structured environment,'' in \emph{Proc. the IEEE Intelligent
  Transportation Systems Conference (ITSC)}, Rhodes, Greece, Sep. 2020.

\end{thebibliography}
\bibliographystyle{IEEEtran}

\end{document}